\documentclass{article}

\usepackage{microtype}
\usepackage{graphicx}
\usepackage{subfigure}
\usepackage{booktabs} 

\usepackage{hyperref}

\usepackage[accepted]{icml2021}

\usepackage{amsmath}
\usepackage{amssymb}
\usepackage{mathtools}
\usepackage{amsthm}
\usepackage{nicefrac}

\renewcommand{\vec}[1]{\mathbf{#1}}
\newcommand{\x}{\ensuremath{\vec{x}}}
\newcommand{\Sspace}{\ensuremath{\mathcal{S}}}
\newcommand{\Xspace}{\ensuremath{\mathcal{X}}}
\newcommand{\emb}[1]{\ensuremath{\operatorname{emb}(#1)}}
\newcommand{\ddG}{\ensuremath{\Delta\Delta G}}

\DeclarePairedDelimiterX{\norm}[1]{\lVert}{\rVert}{#1}

\usepackage[capitalize,noabbrev]{cleveref}
\AtBeginEnvironment{appendix}{\crefalias{section}{appendix}}

\theoremstyle{plain}

\theoremstyle{definition}

\theoremstyle{remark}

\usepackage[textsize=tiny]{todonotes}

\icmltitlerunning{Submission and Formatting Instructions for ICML 2024}

\begin{document}

\twocolumn[
\icmltitle{Active learning for affinity prediction of antibodies}



\icmlsetsymbol{equal}{*}

\begin{icmlauthorlist}
\icmlauthor{Alexandra Gessner}{equal,a}
\icmlauthor{Sebastian W. Ober}{equal,b}
\icmlauthor{Owen Vickery}{equal,b}
\icmlauthor{Dino Ogli\'c}{a}
\icmlauthor{Talip U\c{c}ar}{a}
\end{icmlauthorlist}

\icmlaffiliation{a}{Centre for AI, BioPharmaceuticals R\&D, AstraZeneca}
\icmlaffiliation{b}{Biologics Engineering, Oncology R\&D, AstraZeneca}

\icmlcorrespondingauthor{Alexandra Gessner}{alexandra.gessner@astrazeneca.com}

\icmlkeywords{Machine Learning, ICML}

\vskip 0.3in
]



\printAffiliationsAndNotice{\icmlEqualContribution} 

\begin{abstract}
The primary objective of most lead optimization campaigns is to enhance the binding affinity of ligands. For large molecules such as antibodies, identifying mutations that enhance antibody affinity is particularly challenging due to the combinatorial explosion of potential mutations.
When the structure of the antibody-antigen complex is available, relative binding free energy (RBFE) methods can offer valuable insights into how different mutations will impact the potency and selectivity of a drug candidate, thereby reducing the reliance on costly and time-consuming wet-lab experiments. However, accurately simulating the physics of large molecules is computationally intensive. We present an active learning framework that iteratively proposes promising sequences for simulators to evaluate, thereby accelerating the search for improved binders. 
We explore different modeling approaches to identify the most effective surrogate model for this task, and evaluate our framework both using pre-computed pools of data and in a realistic full-loop setting.

\end{abstract}

\section{Introduction}
\label{sec:intro}

Proteins are complex macromolecules composed of linear chains of amino acids folded into a tertiary structure. Antibodies are noteworthy proteins that interact with a diverse range of molecules via their complementarity-determining regions (CDRs). The immune system utilizes a wide variety ($>10^7$) of antibodies to defend against non-self molecules, with each antibody being able to discriminate between different antigens. When confronted with a potential antigen, the immune system has the remarkable ability to continually refine and optimize the specificity and strength of antibody binding through a process called somatic hypermutation (SHM). Through processes such as affinity maturation, B cells undergo iterative rounds of mutation and selection, leading to the production of antibodies with enhanced binding capabilities against specific antigens. SHM ensures that the immune response evolves to effectively combat a diverse array of antigens and adapt to changing environmental conditions. We can emulate such a biological process by applying active learning to relative binding free energy (RBFE) methods, which model the antibody-antigen interaction to predict their binding affinity.

Active learning is a framework from experimental design that focuses on making informed decisions about which experiments to perform next. This is achieved by constructing a model based on current observations and using the model's predictions to decide which inputs to query in subsequent experiments. This iterative approach, which involves alternating between modeling and acquiring new data points, is data-efficient for improving models, making it particularly useful in scenarios where experiments are costly, such as RBFE simulation.
Bayesian optimization \citep{garnett_bayesian_2023} employs this approach for the global optimization of expensive-to-evaluate black-box functions, making it a suitable method for identifying improved binders using RBFE methods.

The primary goal of this work is the systematic identification of novel antibody mutations that lead to an improved binding affinity to a particular antigen. To this end, we present the following contributions.
\begin{enumerate}
    \item We construct a Bayesian optimization loop that interacts with RBFE methods to find antibody sequences with improved binding affinity.
    \item We introduce a simple yet effective encoding scheme based on the BLOcks SUbstitution Matrix (BLOSUM) \citep{henikoff1992amino, henikoff1992blosum} and evaluate it together with other encoding schemes.
    \item We validate our method on pre-calculated data and evaluate a range of antibody sequence encodings and model choices.
    \item We run the most promising setups from the validation runs in a full loop and find that our algorithm consistently finds vastly improved binders over the mutations available in the pre-computed dataset.
\end{enumerate}

\section{Methods}

\begin{figure}[t]
    \begin{center}
        \includegraphics[width=0.75\columnwidth]{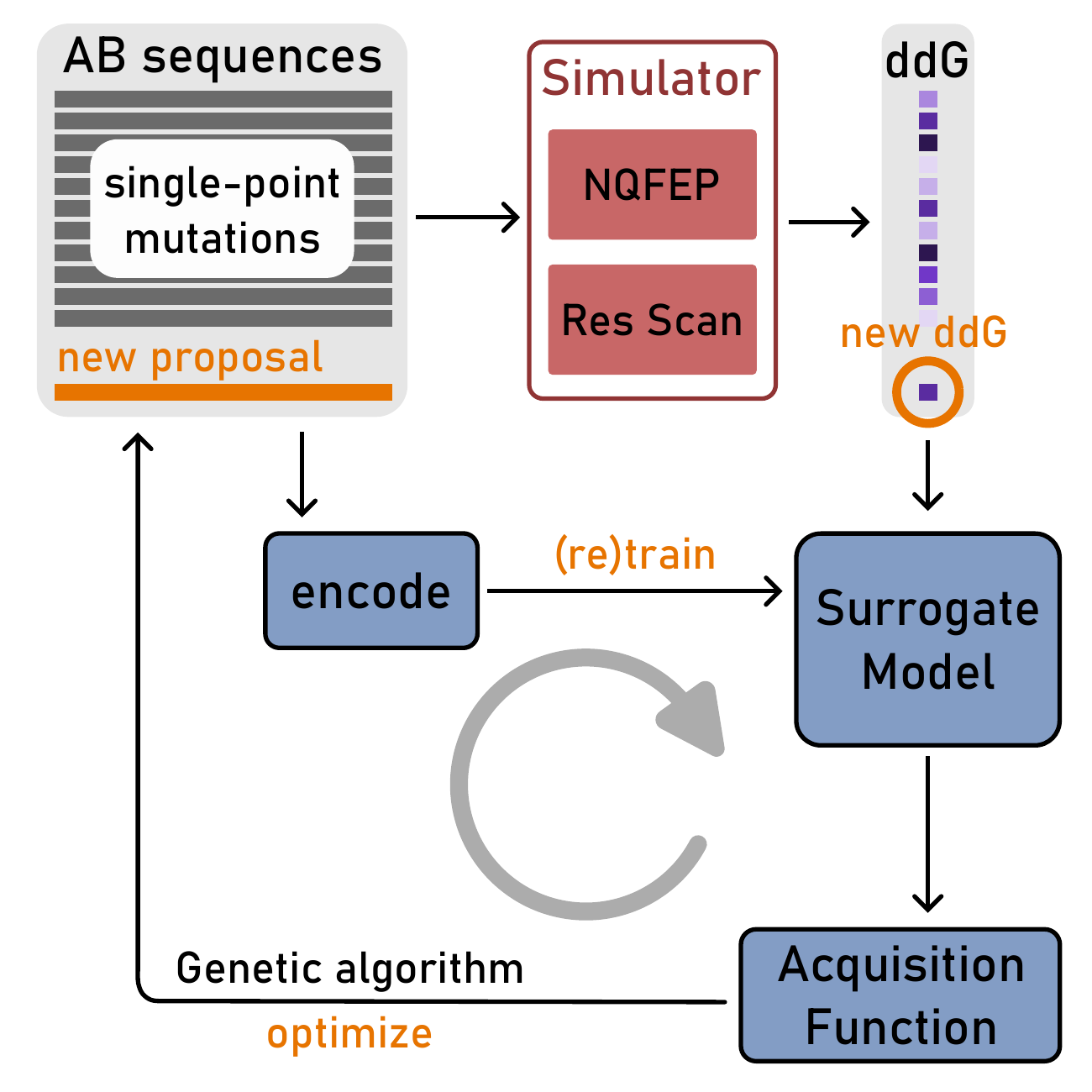}
    \end{center}
    \caption{Schematic of the full loop. In each iteration, a new, previously unseen sequence is proposed by maximizing the acquisition function and fed into the simulator to obtain the corresponding \ddG~value. Both sequence and simulator output are added to the dataset to update the surrogate model.}
    \label{fig:loop_sketch}
\end{figure}

Our full active learning loop consists of the building blocks shown in \cref{fig:loop_sketch}. We initially train the surrogate model with a set of single-point mutations of our wild-type antibody for which we have results from an RBFE simulator (\cref{sec:fep}). We run Bayesian optimization (\cref{sec:bo}) to sequentially propose new sequences that improve binding affinity. The new candidate is proposed by optimizing an acquisition function (\cref{sec:alseq}) and triggers a new query of the simulator. The encoded sequence (\cref{sec:encodings}) and the simulated output are added to the pool and used to update the surrogate model (\cref{sec:gpseq}).

\subsection{Relative binding free energy methods}
\label{sec:fep}
Accurate in-silico prediction of binding affinity has been a significant goal over the past few decades. In particular, relative binding free energy (RBFE) methods are of interest for modeling accurate biological systems, incorporating protein flexibility, explicit solvents, co-factors, and entropic effects. However, these methods have been constrained by high computational costs, limited force field accuracy, and automation challenges, predominantly confining their use to the field of computational chemistry. Moreover, the computational demands, structure prediction, and structural dynamics within biologics hinder large-scale affinity predictions. Despite these challenges, RBFE methods provide valuable insights into the thermodynamics of protein-protein interactions, aiding in the design of biologics with improved binding affinity or specificity.

We employ RBFE simulators to estimate the relative binding affinity \ddG~of an antibody that has been mutated with respect to a fixed parent and target, based on their sequences of amino acids.
The RBFE simulators compute $\ddG=(\Delta G_{\mathrm{bound}})-(\Delta G_{\mathrm{unbound}})$, where $(\Delta G_{\mathrm{bound}})$ and $(\Delta G_{\mathrm{unbound}})$ represent the difference in Gibbs free energy between the wild type (WT) and the mutant in the bound and unbound states respectively. \ddG~hence denotes the gain or loss of energy from forming an antibody-antigen complex relative to the WT, and a negative \ddG~indicates an enhancement in binding affinity.
For this study, we use two different RBFE methods with varying degrees of dynamics and, consequently, different runtimes.

\begin{description}
\item[NQFEP:] Utilizing our own implementation of the non-equilibrium free energy perturbation (NQFEP) method with GROMACS and pmx \citep{seeliger2010protein}, we can calculate the $\Delta G$ of a specific mutation in both the bound and unbound states. By completing two legs of the thermodynamic cycle, we can calculate the $\Delta\Delta G$ of the mutation. The NQFEP method allows for some flexibility in the protein, but it is limited by simulation times on the order of nanoseconds. This simulator provides fairly accurate estimates but comes with a high computational cost of 6-24 hours on a modern GPU.
\item[Schr\"{o}dinger Res Scan:] This is a cheaper but less accurate alternative that estimates the absolute energies of the bound complex and the unbound states, taking the $\ddG$ as the difference between these absolute energies. This method is error-prone, particularly because it uses a single snapshot of the complex. However, this simulation runs in a matter of minutes on a few CPUs.
\end{description}

\subsection{Bayesian optimization}
\label{sec:bo}
Bayesian optimization \citep[BO;][]{garnett_bayesian_2023} is a black-box optimization technique aimed at finding the minimum (without loss of generality) of a function $f : \Xspace \rightarrow \mathbb{R}$, where in our case $\Xspace$ represents the space of antibodies, and $f$ denotes the true $\ddG$.
This technique achieves its goal by using observed input-output pairs $(\x_i, y_i)$ to construct a \emph{surrogate model}.
A crucial aspect of the surrogate model is its ability to estimate the model's uncertainty, which enables an efficient exploration-exploitation trade-off. This trade-off is made concrete by using an acquisition function $a: \Xspace \rightarrow \mathbb{R}$ to select the next antibody to acquire a simulated $\ddG$ value for, which aims to estimate the utility of acquiring a new point by using the predictions (and uncertainty) of the surrogate model.
Following the observation of an initial set of points, we alternate between building a surrogate model and acquiring new points until our budget (e.g., time, or total compute) is exhausted. 

The most common, and one of the most effective, classes of surrogate models for Bayesian optimization are Gaussian processes \citep[GPs;][]{rasmussen2006gaussian}.
We model the observations as $y_i = f(\x_i) + \epsilon_i$, with $\epsilon_i \sim \mathcal{N}(0, \sigma^2)$, placing a GP prior on $f$.
A GP is defined entirely by its mean function, $\mu: \Xspace \rightarrow \mathbb{R}$ (which we restrict to be a learned constant), and its kernel function, $k: \Xspace \times \Xspace \rightarrow \mathbb{R}$, which measures the similarity between two points.

\subsection{Encoding antibody sequences}
\label{sec:encodings}
GPs do not naturally operate on sequence data. In order to use them on antibody sequences, we need to map strings of amino acids $s\in\Sspace$ to numerical values $\x\in\Xspace$, $\emb{s}: s \mapsto \x$. We compare the following embeddings:
\begin{description}
    \item[One-hot encoding:] Each letter in the alphabet of amino acids gets assigned a unique one-hot vector and the encoding is their concatenation according to the sequence. The size of an encoded sequence is therefore the product of sequence length and vocabulary length.
\item[Bag of amino acids:] We count matching $n$-grams of amino acids, corresponding to the bag of words embedding \citep{jurafsky2000speech}. We set $n = 5$. 
    \item[BLOSUM:] The Blocks Substitution Matrix \citep[BLOSUM;][]{henikoff1992blosum} is a substitution matrix used in bioinformatics for sequence alignments of proteins. It is an indefinite similarity matrix quantifying the similarity between pairs of amino acids, computed by using their observed substitutions in related proteins. We perform an eigendecomposition of this kernel matrix, $U D U^{\top}$, and encode the individual amino acids with the rows of $U \vert D\vert^{\nicefrac{1}{2}}$. This is motivated by observations in~\citet{pmlr-v80-oglic18a,pmlr-v97-oglic19a} on flip-spectrum transformations of indefinite kernels.
    \item[AbLang2:] AbLang2 \citep{olsen_ablang2_2024} uses an antibody-specific language model to encode the light and heavy chain of the antibody jointly.
\end{description}

\subsection{Gaussian processes on sequence embeddings}
\label{sec:gpseq}

We consider two types of GP models that are apt to handle the high-dimensional sequence embeddings; firstly, dot-product covariance functions, where we focus on the Tanimoto kernel \citep{ralaivola2005graph},
\begin{equation}
    k_\mathrm{Tanimoto} (\x, \x') = \frac{\langle\x,\x'\rangle}{\norm{\x} + \norm{\x'} - \langle\x,\x'\rangle},
\end{equation}
and secondly, stationary kernels. For the latter, we use the RBF and Mat\'{e}rn-$\nicefrac{3}{2}$ kernels after projecting the embedded data to a lower-dimensional space where these covariance functions are known to work better.
For the dimension reduction we use random matrices of size $N_{\mathrm{emb}} \times N_\mathrm{low}$ with entries sampled from a normal distribution, partly inspired by \citet{wang2016bayesian}.

\subsection{Active learning on sequence data}
\label{sec:alseq}
In order to propose new mutations of the parental antibody for querying the simulator, we require a mechanism for optimizing the acquisition function.
To do so, we follow the work of \citet{moss2020boss}, and use a genetic algorithm, which ``evolves" a ``population" of sequences to maximize the ``fitness" of the sequences, in this case the acquisition value.
The evolution is achieved by a \emph{mutation} operation, which introduces a single amino acid mutation, and a \emph{crossover} operation, which takes two existing sequences, splices them at a random location, and combines the splices to create a new sequence.
In practice, to implement our genetic algorithm, we extend the \texttt{mol\_ga} package \citep{tripp2023genetic} \citep[based off the packages of][]{brown2019guacamol, jensen2019graph} to handle mutation and crossover operations for protein sequences.

\begin{figure}[t]
    \begin{center}
        \includegraphics[width=0.85\columnwidth]{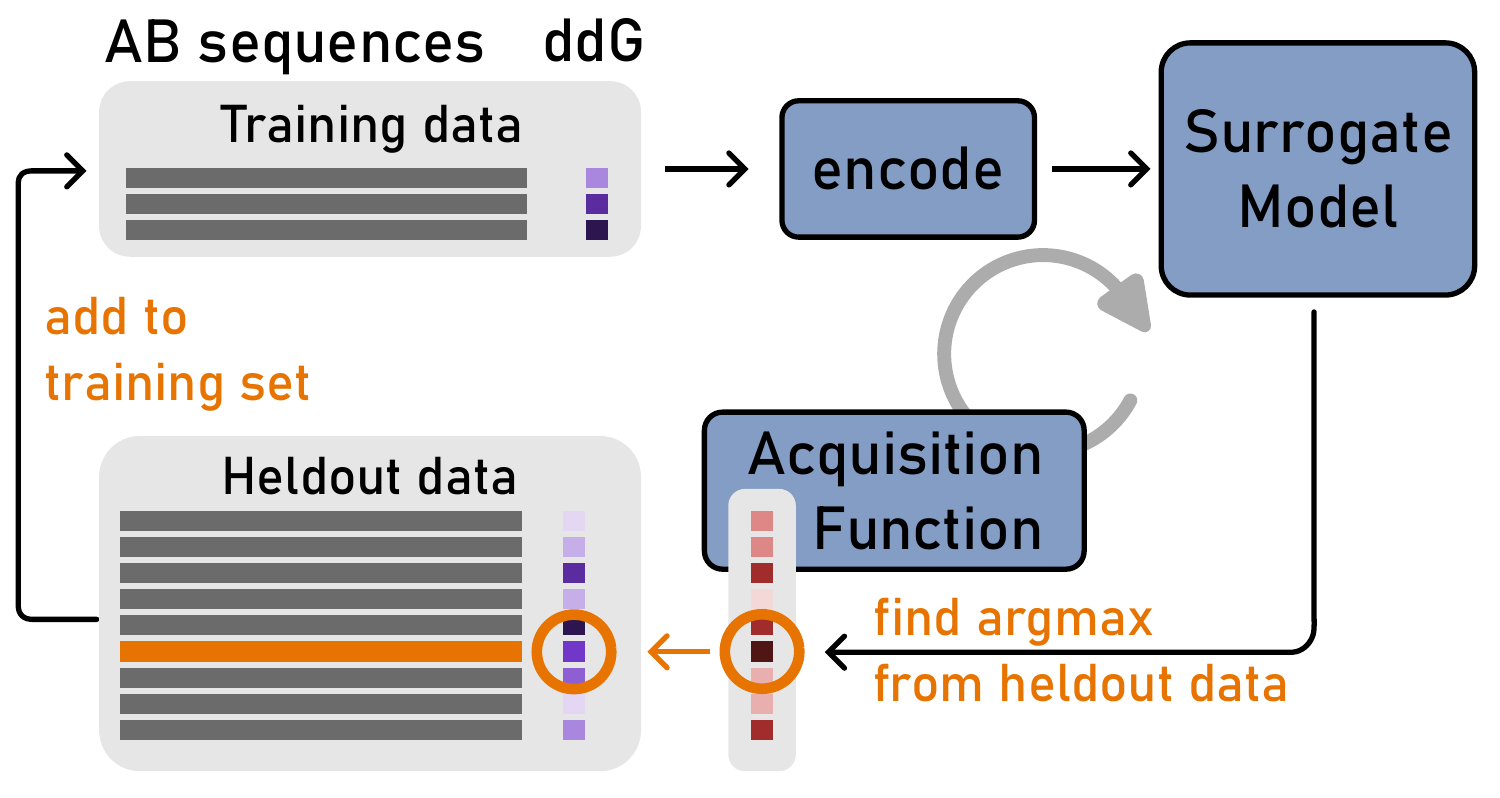}
    \end{center}
    \caption{Schematic of the validation loop. We split the pre-computed dataset into a training set and a pool of held-out data. The loop iteratively selects a new sequence from the pool by maximizing the acquisition function on the held-out data and updates the surrogate model.}
    \label{fig:valloop}
\end{figure}
\section{Data}
\label{sec:data}

We pre-simulated \ddG~values for selected mutations of a parent antibody, including $532$ single-point mutations from the costly NQFEP simulator and $60,479$ mutations from the Schr\"{o}dinger Res Scan simulator (cf. \cref{sec:fep}).

The input data comprises sequences of amino acids for both the light and heavy chains of the antibody. We use the standard single-letter codes representing the 20 amino acids. 
Except for AbLang2, all embeddings concatenate the light and heavy chains using an auxiliary character to treat them as a single sequence. The resulting sequences have a fixed length of 238, obviating the need for padding. With the inclusion of the concatenation character, our alphabet comprises 21 characters. Consequently, the embeddings are of dimensions of 4998 for the one-hot and BLOSUM encodings, 835 (NQFEP) and 834 (Schr\"odinger Res Scan) for the bag-of-amino-acids employing 5-grams, respectively, and 480 for AbLang2.
\begin{figure*}[t]
    \includegraphics[width=\textwidth]{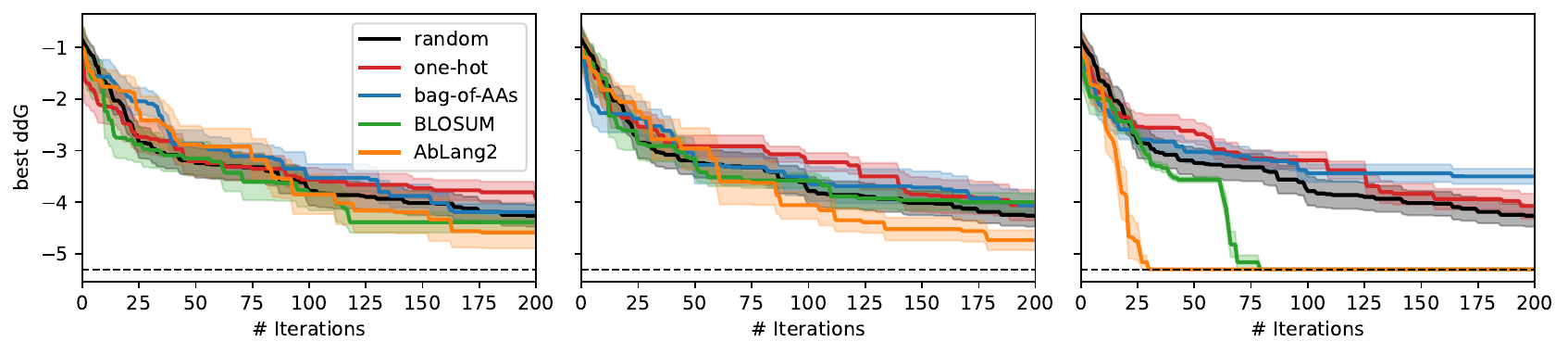}
    \caption{Validation on the NQFEP pre-computed dataset over 200 iterations averaged over 10 runs. Best \ddG~value found using the RBF \emph{(left)}, Mat\'ern \emph{(center)}, and Tanimoto \emph{(right)} kernels, respectively, for all encodings. In the case of the RBF and Mat\'ern kernel, the embeddings have been projected to 5 dimensions. The horizontal dashed line is the best value in the dataset.}
    \label{fig:kabat-fixnoise-5}
\end{figure*}

\section{Experiments}
\label{sec:exp}

Before running the full loop (\cref{fig:loop_sketch}), we validate our model and compare the performance of different encoding schemes (\cref{sec:encodings}) and covariance functions (\cref{sec:gpseq}) to inform experiments using the full loop.

\subsection{Validation}
\label{sec:val_exp}
To validate our method, we run the BO loop on both pre-simulated datasets for different embedding schemes and choices of GP model. We initialize the GP with a small subset of the data, $0.01\%$ ($0.001\%$) of the NQFEP (the Schr\"{o}dinger Res Scan) data.
In each iteration of the Bayesian optimization loop, we evaluate the acquisition function on the heldout data and find the new candidate by choosing the maximizing value. We use expected improvement from \texttt{BoTorch} \citep{balandat2020botorch} as acquisition function and fix the observation variance $\sigma^2$ to $10^{-4}$. Fixing the noise variance is motivated by the determinism of the Schr\"odinger Res Scan simulations and the low variance expected for the NQFEP simulations. Further experiments in which the noise variance is optimized for can be found in \cref{app:addexp}.
For selected choices of covariance function and embedding, we run 10 trials with 200 loop iterations each and track the best value found by the algorithm. Our baseline is a random strategy that records the best \ddG~observed from random picks of a $(\text{sequence}, \ddG)$-pair from the remaining pool. The validation loop is conceptually visualized in \cref{fig:valloop}.

\paragraph{NQFEP data} For the smaller dataset, we ran the validation for all combinations of GP models (\cref{sec:gpseq}) with all encodings (\cref{sec:encodings}). For the latter, we additionally explore the effect of the dimension we project the embedded sequences to prior to feeding them into the kernel. We record the best \ddG~value found by the Bayesian optimization routine up to the current iteration and present the results in \cref{fig:kabat-fixnoise-5} for fixed noise variance.\\
\cref{app:nqfep} contains additional experiments for both fixed and trained noise variance. We observe that the choice of projection dimension has little influence on the performance of the BO model. The choice of encoding appears to be more important where AbLang2 consistently outperforms the other encodings. This is especially true for the Tanimoto kernel (right panel of \cref{fig:kabat-fixnoise-5}), where the best value in the dataset ($-5.3\,\nicefrac{\mathrm{kcal}}{\mathrm{mol}}$) is found within $30$ iterations across all ten trials.

\begin{figure*}[t]
    \includegraphics[width=\textwidth]{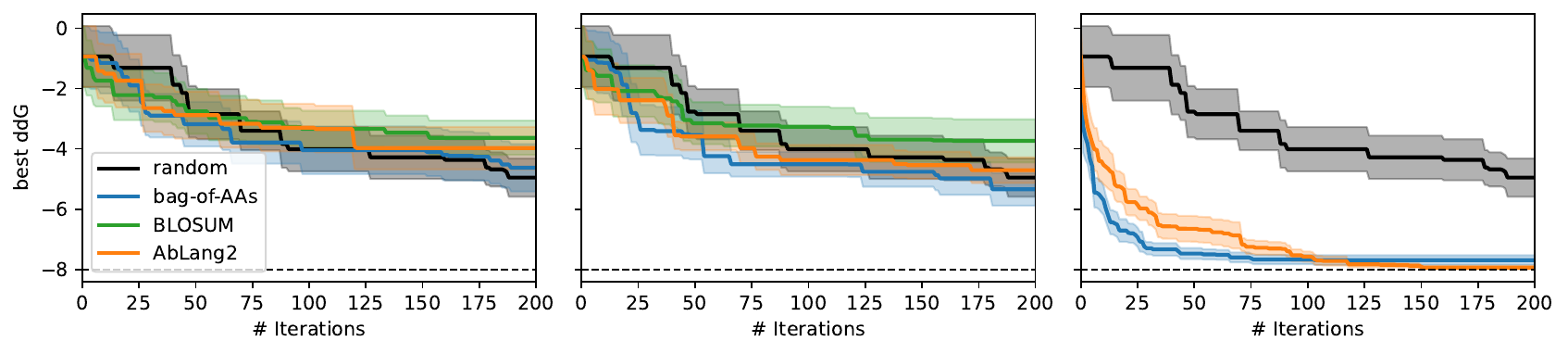}
    \caption{Validation on the Schr\"odinger Res Scan pre-computed dataset over 200 iterations averaged over 10 runs. Best \ddG~value found using the RBF \emph{(left)}, Mat\'ern \emph{(center)}, and Tanimoto \emph{(right)} kernels, respectively, for all encodings. In the case of the RBF and Mat\'ern kernel, the embeddings have been projected to 5 dimensions. The horizontal dashed line is the best value in the dataset.}
    \label{fig:sres-fixnoise}
\end{figure*}

\paragraph{Schr\"odinger Res Scan data} The validation runs are computationally more demanding due to the size of the dataset. We therefore restrict the selection of model settings and compare the Tanimoto and stationary kernels for all embeddings except the one-hot encoding, while fixing the dimension reduction to five. The results for fixed noise variance are displayed in \cref{fig:sres-fixnoise}; those for trained noise variance can be found in \cref{app:sres}. While the stationary kernels do not even outperform the random strategy, the Tanimoto kernel with the AbLang2 encoding finds the best value in the dataset ($-8.00\,\nicefrac{\mathrm{kcal}}{\mathrm{mol}}$) within $\approx150$ iterations. For the Tanimoto kernel, the BLOSUM encoding did not terminate in validation mode due to the unfavorable combination of dataset and embedding size.

\begin{figure}[t]
    \begin{center}
        \includegraphics[width=\columnwidth]{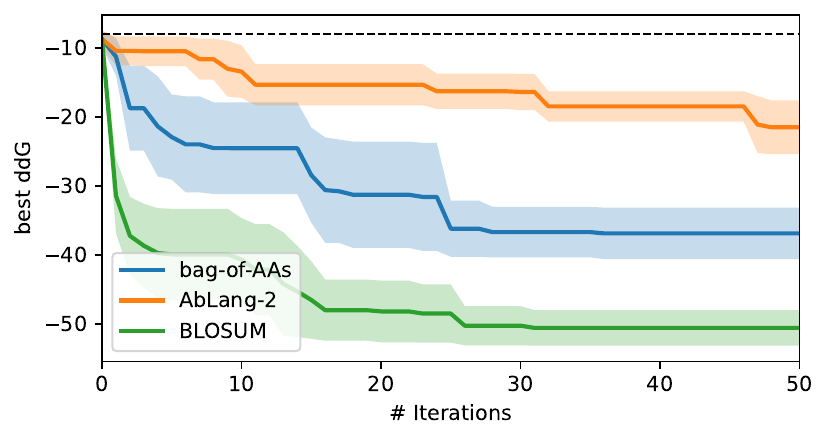}
    \end{center}
    \caption{Results of the full loop run with the Schr\"odinger Res Scan simulator. We plot the best \ddG~values found for each of three encodings using the Tanimoto kernel, as well as the best value from the pooled data from the validation experiments as a dashed line.}
    \label{fig:fullloop}
\end{figure}

\subsection{Full loop}
We use the results from \cref{sec:val_exp} to inform model choices when placing the simulator in the loop. Due to computational constraints, we only consider the Schr\"{o}dinger Res Scan simulator. 
From the validation experiments, we select the AbLang2, BLOSUM, and bag-of-AAs encodings and choose the Tanimoto kernel for the GP.
We initialize the loop using single-residue mutants of the wild type, choosing 3 random mutations at each residue in the antibody's CDR.
We then run the BO loop for 50 iterations, using the GA-based acquisition to select new queries from the set of all possible CDR mutations to send to the simulator.
Each experiment is repeated 3 times.
We present the results of the experiment using the full loop in \cref{fig:fullloop}.
We observe that for all encodings, using an unrestricted search space allows us to quickly find significantly better \ddG~values than using an explicit pool of allowed sequences.
Interestingly, the AbLang2 encoding now shows the weakest performance of all the methods, suggesting that it might be spending more resources exploring the space than the other encodings, as the space is extremely large.

\section{Related work}
\label{sec:related}

\textbf{Bayesian optimization} has been used to optimize sequences in both discrete sequence space \citep{lodhi2002text, moss_henrymossboss_2023} and continuous latent space \citep{stanton2022accelerating, gomez2018automatic}. For example, substring kernels combined with a black-box optimization method such as GA are proposed for optimization in discrete space \citep{moss_henrymossboss_2023, khan2022antbo} while the optimization in latent space is done through embeddings of sequences in a generative modelling setting \citep{grosnit2021high, deshwal2021combining}.

\textbf{Scoring functions} used for optimizing molecules for binding affinity include docking scores \citep{noh2022path, jeon2020autonomous}, relative binding free energy (RBFE) \citep{ghanakota2020combining, moore2023automated} and absolute binding free energy (ABFE) \citep{feng2022absolute, eckmann2024mfbind, eckmann2022limo}, and can be used as a proxy for binding affinity. Moreover, the docking score is known to be inaccurate and has low correlation with experimentally measured affinities \citep{coley2020autonomous, handa2023difficulty, pinzi2019molecular} while scores based on binding free energy tend to be more accurate, but computationally more expensive, than docking scores \citep{moore2023automated}. Since we focus on lead optimization in this work, we use RBFE based simulators as they tend to be computationally more efficient.


\section{Conclusion and Outlook}
\label{sec:conclusion}

Our work attempts to accelerate the in silico search for antibodies with improved binding affinity.
To this end, we constructed an active learning loop over expensive-to-evaluate FEP simulators to efficiently propose antibody sequences that result in improved binders to a specific antigen.
We studied various choices of sequence encodings and model choices for Bayesian optimization in sequence space and found well-performing model choices to be run in the full loop.

There are a number of avenues for future work. Due to the cubic cost of exact GP regression, including all the available data can be prohibitive, potentially requiring inducing points to reduce the computational load. Given multiple simulators of varying cost and accuracy, we plan to extend the model for multi-source Bayesian optimization \citep{poloczek_multi-information_2017} by building a joint probabilistic surrogate model over all simulators. Finally, we expect the active learning scheme to perform better once structural information about the antibody is included in the model.




\bibliography{bib}
\bibliographystyle{icml2024}

\newpage
\appendix
\onecolumn
\section{Additional information}

\subsection{Validation experiments}
\label{app:addexp}

\subsubsection{Additional experiments for NQFEP}
\label{app:nqfep}

\paragraph{Fixed noise variance}

\begin{figure*}[h]
    \includegraphics[width=\textwidth]{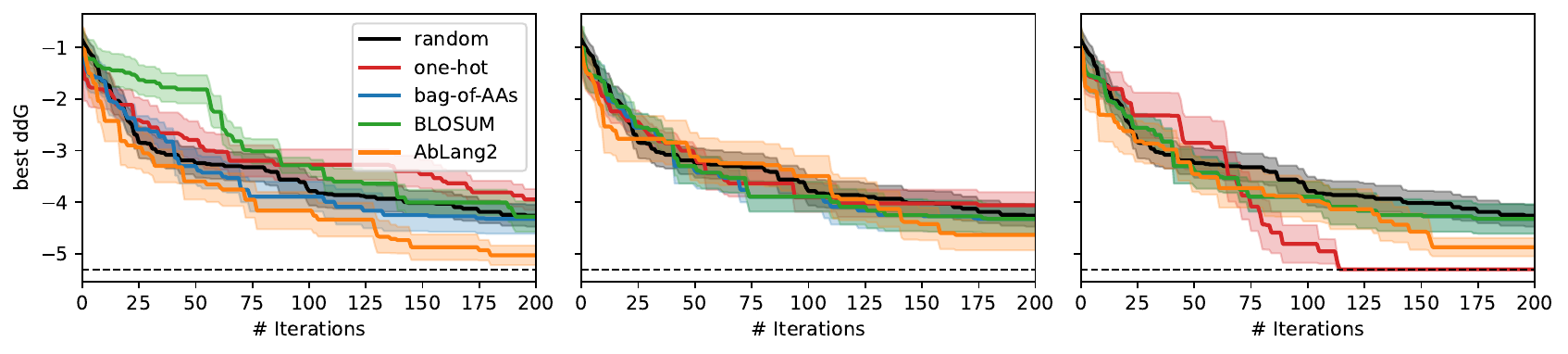}
    \caption{Validation on the NQFEP pre-computed dataset over 200 iterations averaged over 10 runs. Best \ddG~value found using the RBF kernel with inputs projected to 10 \emph{(left)}, 15 \emph{(center)}, and 20 \emph{(right)} dimensions using random projections, respectively, for all encodings. The horizontal dashed line is the best value in the dataset.}
    \label{fig:kabat-fixnoise-rbf}
\end{figure*}

\begin{figure*}[h]
    \includegraphics[width=\textwidth]{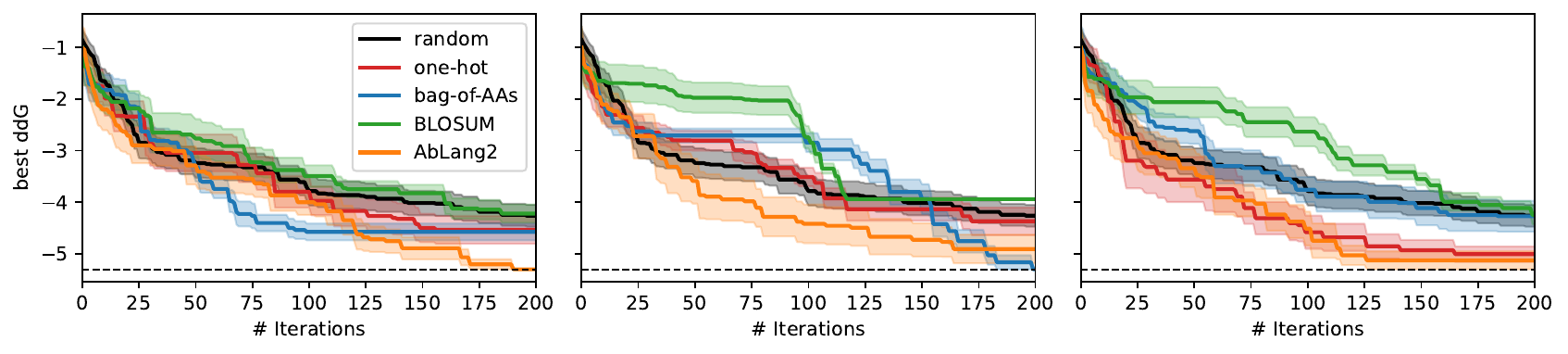}
    \caption{Validation on the NQFEP pre-computed dataset over 200 iterations averaged over 10 runs. Best \ddG~value found using the Mat\'ern kernel with inputs projected to 10 \emph{(left)}, 15 \emph{(center)}, and 20 \emph{(right)} dimensions using random projections, respectively, for all encodings. The horizontal dashed line is the best value in the dataset.}
    \label{fig:kabat-fixnoise-matern}
\end{figure*}

\cref{fig:kabat-fixnoise-rbf} and \cref{fig:kabat-fixnoise-matern} show additional results for the RBF and Mat\'ern kernel, respectively, for different projection dimensions for the embeddings. Only the AbLang2 encoding consistently outperforms the random strategy. At larger projection dimensions, the one-hot encoding also displays good performance.

\begin{figure*}[ht]
    \includegraphics[width=\textwidth]{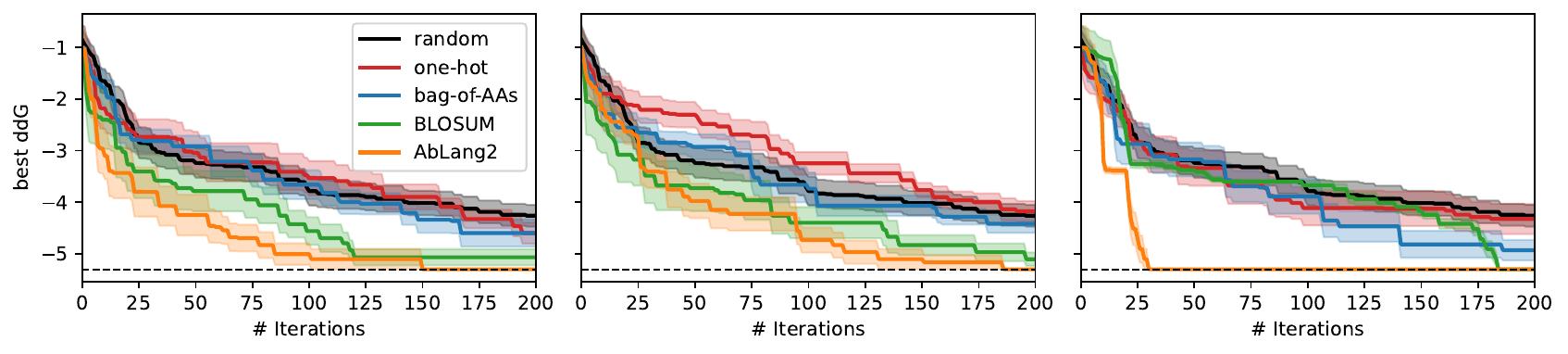}
    \caption{Validation on the NQFEP pre-computed dataset. Same as \cref{fig:kabat-fixnoise-5}, but with learned noise variance,  RBF \emph{(left)}, Mat\'ern \emph{(center)}, and Tanimoto \emph{(right)} kernel.}
    \label{fig:kabat-varnoise-5}
\end{figure*}

\begin{figure*}[ht]
    \includegraphics[width=\textwidth]{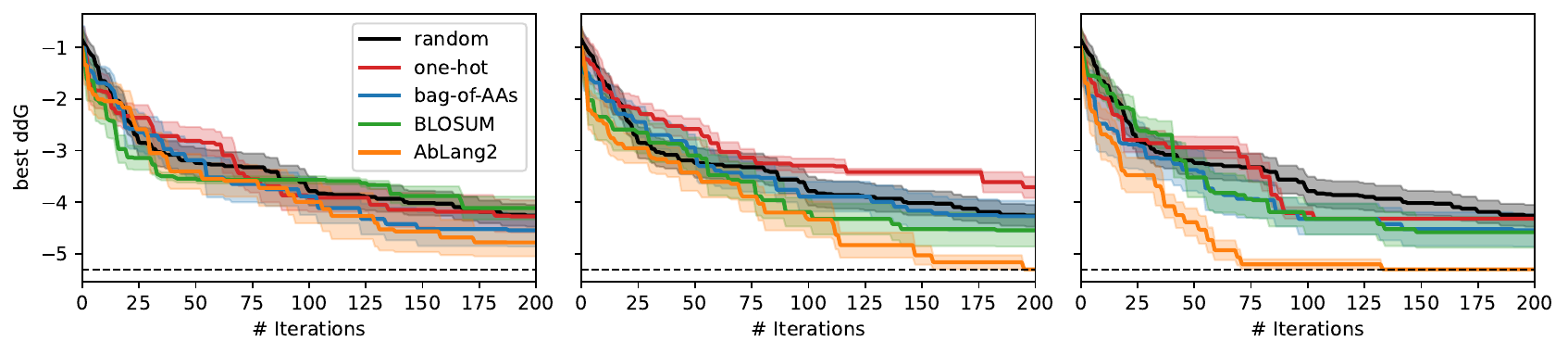}
    \caption{Validation on the NQFEP pre-computed dataset. Same as \cref{fig:kabat-fixnoise-rbf}, but with learned noise variance, for the RBF kernel}
    \label{fig:kabat-varnoise-rbf}
\end{figure*}

\begin{figure}[ht]
    \includegraphics[width=\columnwidth]{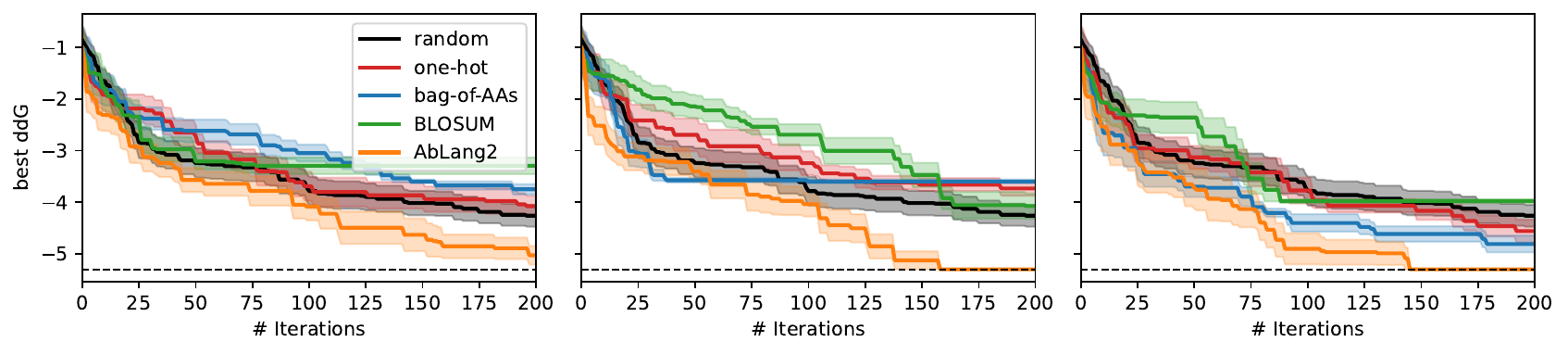}
    \caption{Validation on the NQFEP pre-computed dataset. Same as \cref{fig:kabat-fixnoise-matern}, but with learned noise variance, for the Mat\'ern kernel.}
    \label{fig:kabat-varnoise-matern}
\end{figure}

\paragraph{Trained noise variance}
We present the same results as shown in \cref{sec:val_exp} but train the variance of the likelihood. This requires a modification of the acquisition function in the BO loop, since expected improvement assumes noise-free evaluations. Instead, we employ noisy expected improvement using quasi Monte Carlo \citep{letham2018constrained}, and rely on the implementation in \texttt{BoTorch}.\\
The results in \cref{fig:kabat-varnoise-5}, \cref{fig:kabat-varnoise-rbf}, and \cref{fig:kabat-varnoise-matern} are organized as their counterparts for fixed noise variance in \cref{fig:kabat-fixnoise-5}, \cref{fig:kabat-fixnoise-rbf}, and \cref{fig:kabat-fixnoise-matern}.



\subsubsection{Schr\"odinger Res Scan with learned noise variance}
\label{app:sres}
\begin{figure*}[ht]
    \includegraphics[width=\textwidth]{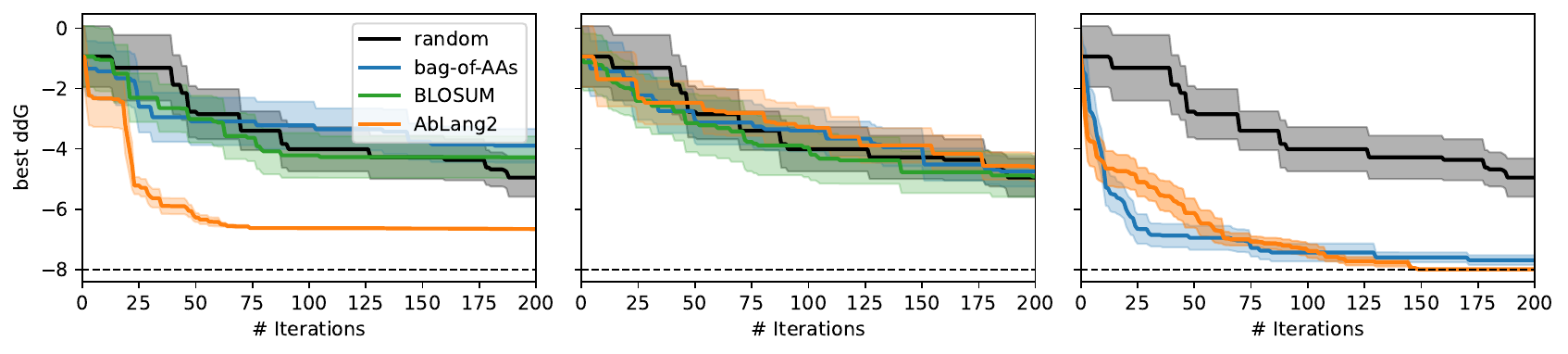}
    \caption{Validation on the Schr\"odinger Res Scan pre-computed dataset. Same as \cref{fig:sres-fixnoise}, but with learned noise variance, for the  RBF \emph{(left)}, Mat\'ern \emph{(center)}, and Tanimoto \emph{(right)} kernel.}
    \label{fig:sres-varnoise}
\end{figure*}

\cref{fig:sres-varnoise} shows the results for the larger dataset when training the variance of the likelihood. Remarkably, the RBF kernel combined with AbLang2 (left panel of \cref{fig:sres-varnoise}) performs significantly better than in the case of fixed noise. However, the training becomes unstable and the runs did not actually complete. This is indicated by the plateau of the curve.
For the Tanimoto kernel, only two encodings, AbLang2 and bag of amino acids, completed its runs. Due to the larger size of the BLOSUM and one-hot encodings, the evaluation of the acquisition function on the entire dataset becomes prohibitively expensive. This is not an issue for the stationary kernels due to the dimension reduction applied on the encoded sequences.

\end{document}